\documentclass{article}

 \PassOptionsToPackage{sort, numbers}{natbib}

 \usepackage[final]{nips_2018}

\usepackage[utf8]{inputenc} 
\usepackage[T1]{fontenc}    
\usepackage{hyperref}       
\usepackage{url}            
\usepackage{booktabs}       
\usepackage{amsfonts}       
\usepackage{nicefrac}       
\usepackage{microtype}      

\usepackage{amsmath}
\usepackage{bbm}
\usepackage{xcolor}
\usepackage{graphicx}
\usepackage[normalem]{ulem}
\useunder{\uline}{\ul}{}
\usepackage{wrapfig} 
\usepackage{enumitem}
\usepackage{xspace}
\usepackage{bm}

\setlist[itemize]{leftmargin=*}
\newcommand{\name}{MAPLE\xspace}

\title{Model Agnostic Supervised Local Explanations}

\author{
  Gregory Plumb\\
  CMU\\
  \texttt{gdplumb@andrew.cmu.edu} \\
  \And
  Denali Molitor \\
  UCLA \\
  \texttt{dmolitor@math.ucla.edu}
  \And
  Ameet Talwalkar \\
  CMU\\
  \texttt{talwalkar@cmu.edu} \\
}

\begin{document}
\maketitle

\begin{abstract}
Model interpretability is an increasingly important component of practical machine learning. 
Some of the most common forms of interpretability systems are example-based, local, and global explanations.  
One of the main challenges in interpretability is designing explanation systems that can capture aspects of each of these explanation types, in order to develop a more thorough understanding of the model.  
We address this challenge in a novel model called \name that uses local linear modeling techniques along with a dual interpretation of random forests (both as a supervised neighborhood approach and  as a feature selection method). 
\name has two fundamental advantages over existing interpretability systems.  
First, while it is effective as a black-box explanation system, \name itself is a highly accurate predictive model that provides faithful self explanations, and thus sidesteps the typical accuracy-interpretability trade-off.
Specifically, we demonstrate, on several UCI datasets, that \name is at least as accurate as random forests and that it produces more faithful local explanations than LIME, a popular interpretability system.  
Second, \name provides both example-based and local explanations and can detect global patterns, which allows it to diagnose limitations in its local explanations.  
\end{abstract}

\section{Introduction}\label{sec:intro}
Leading machine learning models are typically opaque and difficult to interpret, yet they are increasingly being used to make critical decisions: e.g., a doctor's diagnosis (life or death), a biologist's experimental design (time and money), or a lender's loan decision (legal consequences). 
As a result, there is a pressing need to understand these models to ensure that they are correct, fair, unbiased, and/or ethical. 
Although there is no precise definition of interpretability and user requirements are generally application-specific, three of the most common types of model explanations are:
\begin{enumerate}[leftmargin=*]
\item \textbf{Example-based}. 
In the context of an individual prediction, it is natural to ask: \emph{Which points in the training set most closely resemble a test point or influenced the prediction?} 
Nearest neighbors and influence function based methods are archetypal methods that naturally lead to example-based explanations~\cite{bien2011prototype,koh2017understanding,baehrens2010explain}. 
\item \textbf{Local}. 
Alternatively, we may aim to understand an individual prediction by asking: \emph{If the input is changed slightly, how does the model's prediction change?}
Local explanations are typically derived from a model directly (e.g., sparse linear models), or from a local model that approximates the predictive model well in a neighborhood around a specific point~\cite{ribeiro2016should,baehrens2010explain}.
\item \textbf{Global}. 
To gain an understanding of a model's overall behavior we can ask: \emph{What are the patterns underlying the model's behavior?} 
Global explanations usually take the form of a series of rules~\cite{ribeiro2018anchors,lakkaraju2016interpretable}.
\end{enumerate}

\begin{figure}
\centering
\includegraphics[width=.30\textwidth]{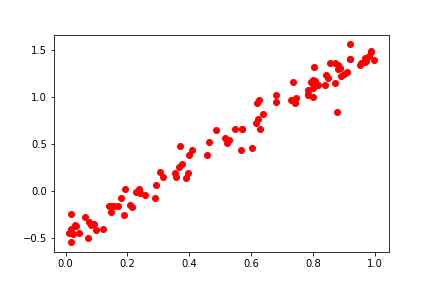}
\includegraphics[width=.30\textwidth]{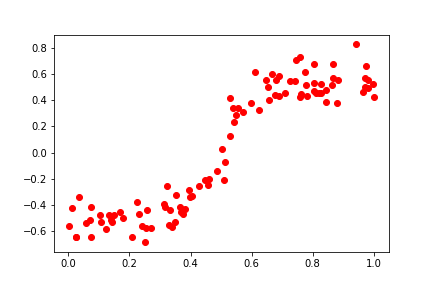}
\includegraphics[width=.30\textwidth]{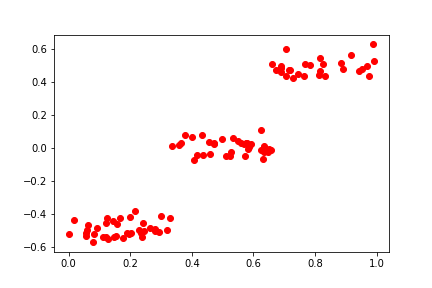}
\caption{\label{fig:3datasets} Toy datasets. (a) Linear. (b) Shifted Inverse Logistic (SIL). (c) Step Function.}
\end{figure}

Example-based explanations are clearly distinct from the other two explanation types, as the former relies on sample data points and the latter two on features.  
Further, local and global explanations themselves capture fundamentally different characteristics of the predictive model. 
To see this, consider the toy datasets in Fig.~\ref{fig:3datasets} generated from three univariate functions.  

Generally, local explanations are better suited for modeling smooth continuous effects (Fig.~\ref{fig:3datasets}a). 
For discontinuous effects (Fig.~\ref{fig:3datasets}c) or effects that are very strong in a small region (Fig.~\ref{fig:3datasets}b), which can be approximated well by discontinuities, they either fail to detect the effect or make unusual predictions, depending on how the local neighborhood is defined. 
We will call such effects \textit{global patterns} because they are difficult to detect or model with local explanations.  
Conversely, global explanations are better suited for global patterns because these discontinuities create natural `rules,' and are less effective at explaining continuous effects because explanation rules must introduce arbitrary feature discretization/binning.  
Most real datasets have both continuous and discontinuous effects and, therefore, it is crucial to devise explanation systems that can capture, or are at least aware of, both types of effects.

In this work, we propose a novel model explanation system that draws on ideas from example-based, local, and global explanations. 
Our method, called \name \footnote{The source code for our model and experiments is at  https://github.com/GDPlumb/\name.}, is a \emph{supervised neighborhood} approach that combines ideas from local linear models and ensembles of decision trees. 
One of the distinguishing features of \name is that, for a particular prediction, it assigns weights to each training point;  these weights induce a probability distribution over the input space that we call the \emph{local training distribution}. 
\name is endowed with several favorable properties described below:
\begin{itemize}[leftmargin=*]
\item It avoids the typical trade-off between model accuracy and model interpretability \cite{lipton2016mythos, doshi2017towards,lundberg2017unified}, as it is a highly accurate predictive model (on par with leading tree ensembles) that simultaneously provides example-based and local explanations.

\item While it does not provide global explanations, it does detect global patterns by leveraging its local training distributions, thus distinguishing it from other local explanation methods. 
As a consequence, it can \emph{diagnose} limitations of its local explanations in the presence of global patterns. 
Moreover, it offers global feature selection, which can help detect data leakage \cite{kaufman2012leakage} or possible bias \cite{ribeiro2018anchors}.  

\item In some settings, we want to represent a predictive model with a small number of exemplar explanations \cite{ribeiro2016should} and, when asked to explain a new test point, must decide which of these exemplar explanations to use.  
Its local training distribution allows us to make that decision in a principled fashion, thereby addressing another key weakness of existing local explanation systems~\cite{ribeiro2018anchors}.  

\item In addition to providing faithful self-explanations, it can also be deployed effectively as a black-box explainer by simply training it on the predictions of a black-box predictive model instead of the actual labels.
We find that it produces more faithful explanations than LIME \cite{ribeiro2016should}, a commonly used model-agnostic algorithm for generating local explanations.  
\end{itemize}  

\section{Background and Related Work} \label{background}

We divide the background material into two sections.  
The first outlines interpretability and the second reviews the random forest literature that is most relevant to \name.  

\subsection{Interpretability} \label{background-int}

We say that a function is \textbf{interpretable} if it is human simulatable \cite{lipton2016mythos}.  
This definition requires that a person can 1) carry out all of the model calculations in reasonable time, which rules out functions that are too complicated, and 2) provide a semantic description of each calculation, which rules out using features that do not have well understood meanings.  
Generally, sparse-linear models, small decision trees, nearest neighbors, and short rule lists are all considered to be human simulatable.

Based on the ideas in \cite{ribeiro2016should,lipton2016mythos}, we define a \textbf{local explanation} at $x$, denoted $exp_x()$, as an interpretable function that approximates $pred()$ well for $x'$ in a neighborhood of $x$ where $pred()$ is the predictive model being explained.
There are two main challenges when using this type of explanation.  
The first is accurately modeling or detecting global patterns.  
Intuitively, local explanations based on supervised neighborhood methods will fail to detect global patterns and those based on unsupervised neighborhoods will be inaccurate near them because their neighborhoods are, respectively, either partitioned on the discontinuity or contain points on both sides of it.
More details about and a visualization of this are in Sec. \ref{training-points}.  
The second challenge is determining if an explanation generated at one point can be applied at a new point \cite{ribeiro2018anchors}. 
We demonstrate how \name addresses these challenges in Sec. \ref{ltd}. 

The most closely related work to ours is LIME \cite{ribeiro2016should}, which provides a local explanation by fitting a sparse linear model to the predictive model's response via sampling randomly around the point being explained.  
The authors of \cite{ribeiro2016should} also demonstrate how to use local explanations in a variety of practical applications.
They also define SP-LIME which summarizes a model by finding a set of points whose explanations (generated by LIME) are diverse in their selected features and their dependence on those features.  
However, this does not address the difficulties local explanations have with global patterns because the individual explanations being chosen all are unaware of them.  

Fundamentally, the problem of identifying a good local explanation is a causal question, which is typically very difficult to answer since most models are not causal.  
However, the local explanation is not trying to find causal structure in the data, but in the model's response.  
This makes the problem feasible because we can freely manipulate the input and observe how the model's response changes.  
However, most explanation systems are not evaluated in a way that is consistent with this goal and use the standard evaluation metric: $\mathbb{E}_x[loss(exp_x(x), pred(x))]$.  
To address this issue, we define the \textbf{causal local explanation metric} as 
\begin{equation}\label{eq:causal}
\mathbb{E}_{x, x' \sim p_x}[loss(exp_x(x'), pred(x'))].
\end{equation}
This metric is based on sampling $x'$ from $p_x$, which is a distribution centered around $x$, and encourages the explanation generated at $x$ to accurately predict the model's value at $x'$.
Interestingly,  performing well on the standard metric does not guarantee doing well on the casual metric.  
To see this, consider a local linear explanation with the form $exp_x(x') = 0^T x' + pred(x)$.  
While it performs perfectly on the standard evaluation metric, its performance on the causal metric can be arbitrarily bad.  
This is because all of the active effects (the features that, if perturbed, significantly effect the model's response) are rolled into the bias term $pred(x)$ rather than being given a non-zero coefficient. 
This explanation does not tell us anything about why the model made its prediction and, consequently, we want a metric that does not select it.  

Based on the methods in \cite{lakkaraju2016interpretable,ribeiro2018anchors}, we define a \textbf{global explanation} as a set of rules that generally hold true for $pred()$ for all or a well defined subset of the input space.  
Two of the main challenges of using global explanations are 1) adequately covering the input space and 2) properly processing the data to allow the rules to be meaningful. 
Anchors \cite{ribeiro2018anchors}, which approximates the model with a set of if-then rules, is an example of a method that provides global explanations.  
It has the advantage of providing easily understood explanations and it is simple to determine if an explanation applies for a specific instance.  
Further, \cite{ribeiro2018anchors} shows that it is possible to choose these rules such that they have very high precision (at the cost of reduced coverage).  
While \name does not directly offer global explanations,  we show how to use it to detect global patterns in Sec. \ref{dgp}, which are the patterns global explanations represent best.   

Based on \cite{lipton2016mythos,koh2017understanding,baehrens2010explain}, we define an \textbf{example-based explanation} as any function that assigns weights to the training points based on how much influence they have on the predictive model or on an individual prediction.  
The most general form of an example-based explanation is influence functions \cite{cook1980characterizations}.  
Influence functions study how the model or prediction would change if a training point was up-weighted infinitesimally, but require model differentiability and convexity to work.  
However, it was recently shown that these assumptions can be relaxed and that influence functions can be used for understanding model behavior, model debugging, detecting dataset errors, and creating visually indistinguishable adversarial training examples \cite{koh2017understanding}.  
More specific strategies for methods of generating example-based explanations are Case Based Reasoning \cite{bien2011prototype}, $K$ Nearest Neighbors \cite{baehrens2010explain}, and SP-LIME \cite{ribeiro2016should}.  
Notably, \name naturally provides influence functions via the weights it assigns to the training points for a particular test-point/prediction.  

\subsection{Random Forests: Feature Selection and Local Models}

Due to their accuracy and robustness, random forests \cite{breiman2001random} have been a popular and effective method in machine learning.  
Unfortunately, they are not generally considered to be interpretable because they aggregate many decision trees, each of which is often quite large.  
However, they yield a measure of global variable importance and they can be viewed as a way of doing supervised neighborhood selection. 
We use both of these aspects in defining our method.  

The permutation based importance measure initially proposed in \cite{breiman2001random} determines the feature importance by considering the performance of the random forest before and after a random permutation of a predictor.  
Another popular variant was proposed in \cite{friedman2001greedy} which works by summing the impurity reductions over each node in the tree where a split was made on that variable, while adjusting for the number of points in the node, and then averaging this over the forest. 
DStump \cite{kazemitabar2017variable} simplifies this measure further by only considering the splits made on the root nodes of the trees.  
We use DStump for global feature selection as part of \name, though \name can be extended to work with other variants.  

One of the reasons that local methods are not commonly applied on large-scale problems is that, although their learning rates are minimax optimal, this rate is conservative when not all of the features are involved in the response, as demonstrated empirically in \cite{bloniarz2016supervised}.
As a result, we are interested in using a supervised local method. 
Random forests have two main interpretations towards this: first, as an adaptive method for finding potential nearest neighbors \cite{lin2006random} and, second, as a kernel method \cite{scornet2016random}.  
Recently, \cite{bloniarz2016supervised} introduced SILO which explicitly uses random forests to define the instance weights for local linear modeling.  
Empirically, they found that this decreased the bias of the random forest and increased its variance, which is potentially problematic in high dimensional settings.

\section{\name}

Our proposed method, \name (\textbf{M}odel \textbf{A}gnostic Su\textbf{P}ervised \textbf{L}ocal \textbf{E}xplanations), combines the idea of using random forests as a method for supervised neighborhood selection for local linear modeling, introduced in \cite{bloniarz2016supervised} as SILO, with the feature selection method proposed in \cite{kazemitabar2017variable} as DStump. 
For a given point, SILO defines a local neighborhood by assigning a weight to each training point based on how frequently that training point appears in the same leaf node as the given point across the trees in the random forest.   
DStump defines the importance of a feature based on how much it reduces the impurity of the label when it is split on at the root of the trees in the random forest. 
These methods are explained in more detail shortly.  

Under certain regularity conditions, SILO has been shown to be consistent in that its estimator converges in probability to the true function \cite{bloniarz2016supervised}.  
Further, DSTump has been shown to identify the active features in a high dimensional setting under the assumption of a general additive model \cite{kazemitabar2017variable}.  
As a result, the combination of these methods is likely an effective model in a variety of problem settings because it should inherit both of these properties.  
However, rather than focus on the statistical properties and empirical accuracy of this combination, we focus instead on how the resulting algorithm can be applied to problems where interpretability is a concern.

Before formally defining these procedures and our method, we introduce some notation. 
Let $x \in \mathbb{R}^{p+1}$ be a feature vector; we assume $[x]_0 = 1$ is a constant term.  
The index $j \in \{0, \ldots, p\}$ will refer to a specific feature in this vector.  
Next, let $\{x_i\}_{i=1}^n$ be the training set.  
The index $i$ will refer to a specific feature vector (training point) in the training set.   
Then $X \in \mathbb{R}^{n \times (p + 1)}$ will denote the matrix representation of the training set (i.e., $[X]_{i,j} = [x_i]_j$).  
Finally, let $\{T_k\}_{k=1}^K$ be the trees in the random forest; the index $k$ will always refer to a specific tree.  

We begin by defining the process by which SILO computes the weights of the training points (i.e., the \emph{local training distribution}) and makes predictions.  
Let $leaf_k(x)$ be the index of the leaf node of $T_k$ that contains $x$.  
We then define the connection function of the $k^{th}$ tree as $$c_k(x, x') = \mathbbm{1}\{leaf_k(x) = leaf_k(x')\}$$ so the number of training points in the same leaf node as $x$ is $$num_k(x) = \sum\limits_{i=1}^n c_k(x_i, x)$$.
Finally, the weight function of the random forest for the $i^{th}$ training point at the point $x$ is 
\begin{equation} \label{weight}
w(x_i, x) = \frac{1}{K} \sum\limits_{k = 1}^{K} \frac{c_k(x_i, x)}{num_k(x)}.
\end{equation}

For a random forest, the model prediction can be written as 
\begin{equation}
\hat{f}_{RF}(x) = \frac{1}{K} \sum\limits_{k = 1}^K \frac{\sum\limits_{i = 1}^n c_k(x_i, x) y_i}{num_k(x)}.
\end{equation}
For SILO, the prediction is given by evaluating the solution to the weighted linear regression problem defined by $\{x_i, w(x_i, x), y_i\}$ at $x$.
Let $W_x \in \mathbb{R}^{n \times n}$ be the diagonal weight matrix where $[W_x]_{i,i} = w(x_i, x)$.  
Then SILO's prediction is 
\begin{equation}
\hat{f}_{SILO}(x) = \hat{\beta_x}^T x \text{ where } \hat{\beta_x} = (X^T W_x X)^{-1}X^T W_x y.
\end{equation}

Next, we define the process DStump uses to select features.
Let $split_k \in \{1, \ldots, p\}$ be the index of the feature that the root node of the $k^{th}$ tree split on and suppose that that split reduces the impurity of the label by $r_k$. 
Then DStump assigns feature $j$ the score $$s_j = \sum\limits_{k = 1}^K \mathbbm{1}\{split_k = j\} r_k$$, and chooses the subset, $A_d \subset \{1, \ldots, p\}$, of the $d$ highest scored features. 

\name combines these procedures by using SILO's local training distribution and the best $d$ features from DStump (along with a constant term corresponding to the bias in the local linear model) to solve the weighted linear regression problem $\{[x_i]_{A_d}, w(x_i, x), y_i\}$.  
Formally, let $Z_d = [X]_{:,{A_d}} \in \mathbb{R}^{n \times (d + 1)}$ and $z_d = [x]_{A_d} \in \mathbb{R}^{d + 1}$.  
Then \name makes the prediction
\begin{equation}
\hat{f}_{\name}(x) = \hat{\beta}_{x,d}^T z_d \text{ where } \hat{\beta}_{x,d} = (Z_d^T W_x Z_d)^{-1} Z_d^T W_x y.
\end{equation}

\textbf{Choosing $d$}:
We pick the number of features to use in our local linear model via a greedy forward selection procedure that relies on the fact that features can be sorted by their $s_j$ scores. 
 Specifically, we evaluate the predictive accuracy of \name on a held out validation set for $d = 1, \ldots, p$ and choose the value of $d$ that gives the best validation accuracy. 
It would also be possible to choose $d$ based on the causal local metric.

\textbf{Extension to Gradient Boosted Regression Trees (GBRT)}:
GBRTs \cite{friedman2001greedy}, an alternative tree ensemble approach, can also naturally be integrated with \name.  Indeed, GBRT tree ensembles can be used to generate local training distributions and feature scores, which can then be fed into \name.

\section{\name as an Explanation System} \label{ltd}

In this section we describe how to use \name to generate explanations (\cite{ribeiro2016should} gives details on practical applications of these explanations).
The general process of using \name to explain a prediction is essentially the same whether we use \name as a predictive model or only as a black-box explainer; the only difference is that in the first case  we fit \name directly on the response variable, while  in the second case we fit it on the predictive model's predicted response.  
\name's local training distribution is vital to these explanations, and is what enables it to address two core weaknesses of local explanations related to 1) diagnosing their limitations in the presence of global patterns, and 2) selecting an appropriate explanation for a new test point when restricted to an existing set of exemplar explanations. 
We discuss both of these topics in the remainder of this section. 

\subsection{Generating Explanations and Detecting Global Patterns} \label{dgp}

When \name makes a prediction/local explanation, it uses a local linear model, where the coefficients determine the estimated local effect of each feature.  
If a feature coefficient is non-zero, then we can interpret the impact of the feature according to the sign and magnitude of the coefficient.  
However, a zero coefficient has two possible interpretations: 1) the feature does not contain global patterns and so it is, indeed, locally inactive; or 2) the feature does contain global patterns and this feature is in fact significant (though not necessarily locally significant). 
Consequently, our main goal is \emph{diagnosing} the efficacy of a local explanation to determine whether a feature with a zero coefficient contains global patterns.  
We summarize how we can leverage the local training distribution to do this (see additional discussion in Sec.~\ref{training-points}).  

\begin{figure}
\centering
\includegraphics[width=0.475\textwidth]{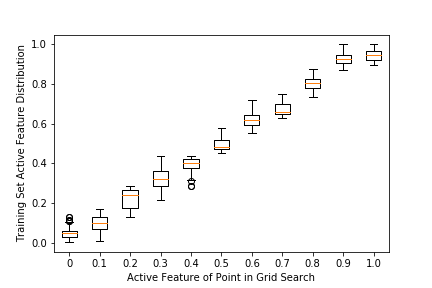}
\includegraphics[width=0.475\textwidth]{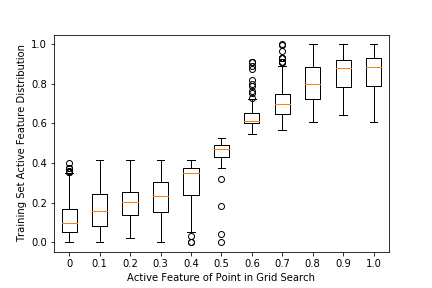}
\includegraphics[width=0.475\textwidth]{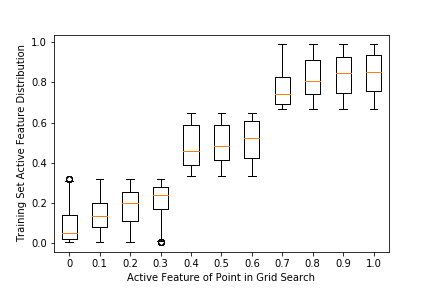}
\caption{\label{fig:distributions} \small 
Distribution of the active feature of the most influential points in the training set across a grid search over the range of feature values.
(a) \textbf{Linear Dataset}: 
The distributions are roughly uniform in width, typically centered, and change smoothly. 
(b) \textbf{SIL Dataset}:
The distributions are wider in the flatter regions of the function and narrower in the transition.  
They also follow two disjoint ranges with one intermediate range on the transition between the two flatter regions.
(c) \textbf{Step Dataset}: 
The distributions are wide, essentially disjoint across the discontinuities of the step function, and not centered. }
\end{figure}

In particular, we propose two diagnostics for each feature. The first (and simpler) diagnostic involves using the local training distribution for the given test point to create a boxplot to visualize the distribution of each feature.
If the boxplot is substantially skewed (i.e., not centered around the test point), then that feature likely contains a global pattern and the test point is nearby it.  
If the boxplots are not skewed, then the second diagnostic involves performing a grid search across the range of the feature in question.  
For each value on the grid, we can sample the remaining features in some reasonable way (e.g., by finding several training points with a similar feature value, assuming feature independence and sampling from the empirical distribution, or via MCMC), and create a boxplot for the local training distribution across this grid (see Fig. \ref{fig:distributions} for examples of this type of plot, the experimental setup is described in Sec. \ref{training-points}).  
If the local training distributions appear to share similar boundaries that change abruptly during the grid search, as seen in Fig. \ref{fig:distributions}c and somewhat in Fig. \ref{fig:distributions}b, then there is likely a global pattern present in that feature.  
Conversely, if the local training distributions are roughly centered around the test point during the grid search and change smoothly during it, as seen in Fig. \ref{fig:distributions}a, then the effect of that feature likely does not have significant global patterns.  

\subsection{Picking an Exemplar Explanation}

As noted in~\cite{ribeiro2016should}, there are settings where we want to compile a small set of (presumably diverse) representative exemplar explanations, and use these exemplar explanations to explain new test points. 
Selecting the appropriate exemplar for a new test point is a challenge for existing local explanation systems~\cite{ribeiro2018anchors}. 
\name provides an elegant solution to this problem by using the local training distributions.  
Specifically, we can determine if we can apply a particular exemplar explanation to a proposed test point by evaluating how likely the proposed point is under the exemplar explanation's local training distribution.  

Of course, there is no guarantee that, collectively, these distributions span the entire input space.  
If asked to explain a test point from an uncovered part of the input space, then the proposed point will have low probability under all of the exemplar explanation distributions, and, having noticed that, we can determine that no exemplar explanation should be applied.  
Similarly, we can detect if multiple exemplar explanations may be equally applicable.  

\section{Experimental Results}

We present experiments demonstrating that:
1) \name is generally at least as accurate as random forests, GBRT, and SILO 
2) \name provides faithful self-explanations, i.e., its local linear model at $x$ is a good local explanation of the prediction at $x$ 
3) \name is more accurate in predicting a black-box predictive model's response than a comparable and popular explanation system, LIME \cite{ribeiro2016should} 
4) The local training distribution can be used to detect the presence of global patterns in the predictive model. 

\begin{table}[tbp]
\centering

\resizebox{\columnwidth}{!}{%
\begin{tabular}{|c|c|ccc|ccc|}
\hline
\textbf{Dataset} & \textbf{LM} & \textbf{RF} & \textbf{SILO + RF} & \textbf{\name + RF}      & \textbf{GBRT}  & \textbf{SILO + GBRT} & \textbf{\name + GBRT}    \\ \hline
Autompgs         & 0.446     & 0.4164    & 0.3784         & {\ul 0.381}          & 0.392      & 0.3745         & 0.377                \\
Communities      & 0.781    & 0.745    & 0.724         & {\ul \textbf{0.688}} & 0.709      & 0.751           & \textbf{0.712}       \\
Crimes           & 0.327    & 1.012    & 0.531          & {\ul \textbf{0.331}} & 0.968     & 0.493             & {\ul \textbf{0.295}} \\
Day              & 0           & 0.204    & 1.7e-05            & {\ul \textbf{6e-06}}    & 0.104      & 1.3e-05              & {\ul \textbf{4e-06}}    \\
Happiness        & 0.001    & 0.644   & 0.001          & {\ul 0.001}          & 0.344       & 0.001             & {\ul 0.001}          \\
Housing          & 0.56    & 0.486   & 0.409          & {\ul 0.419}          & 0.395       & 0.396           & 0.404                \\
Music            & 0.935     & 0.742    & 0.881           & \textbf{0.764}       & {\ul 0.658} & 0.901            & \textbf{0.849}       \\
Winequality-red  & 0.814    & 0.78     & 0.779           & 0.778              & 0.783       & 0.786            & 0.779             \\ \hline
\end{tabular}}
\caption{\label{uci-accuracy} \small
Average RMSE across 50 trials; underlined results indicate that \name differed significantly from the baseline method (RF or GBRT) and bold results indicate that \name differed significantly from SILO built on the same baseline.  
With the exception of the Music dataset, \name is at least as good as the baseline.}

\begin{minipage}{.48\linewidth}
\centering

\begin{tabular}{|c|cc|}
\hline
\textbf{Dataset} & \textbf{LIME} & \textbf{\name} \\ \hline
Autompgs         & 0.178                   & \textbf{0.042}          \\
Communities      & 0.409                   & \textbf{0.130}          \\
Crimes           & 0.276                   & \textbf{0.047}          \\
Day              & 0.034                   & \textbf{0}                \\
Happiness        & 0.05                   & \textbf{3e-05}            \\
Housing          & 0.238                   & \textbf{0.07}          \\
Music            & 0.189                   & 0.181                   \\
Winequality-red  & 0.149                   & \textbf{0.06}           \\ \hline
\end{tabular}\caption{\label{as-self} \small
A comparison of the causal metric of \name vs LIME for $\sigma = 0.1$ when being used to explain the predictions of \name.  
Values shown are RMSE averaged over 25 trials.  
Bold entries denote a significant difference.  }

\end{minipage}%
\hspace{4pt}
\begin{minipage}{.48\linewidth}
\centering

\begin{tabular}{|c|c|cc|}
\hline
\textbf{Dataset} & \textbf{SVR} & \textbf{LIME} & \textbf{\name}  \\ \hline 
Autompgs         & 0.39         & 0.282               & \textbf{0.15}       \\ 
Communities      & 0.761        & 0.338               & \textbf{0.323}      \\ 
Crimes          & 0.895        & 0.183               & 0.232               \\
Day              & 0.2          & 0.242               & \textbf{0.17}       \\ 
Happiness        & 0.267        & 0.28                & \textbf{0.187}      \\  
Housing          & 0.459        & 0.366               & \textbf{0.206}      \\
Music            & 0.816        & 0.326               & \textbf{0.304}      \\ 
Winequality-red  & 0.807        & 0.295               & \textbf{0.204}      \\ \hline 
\end{tabular}
\caption{\label{as-bb} \small
A comparison of the causal metric of \name vs LIME for $\sigma = 0.1$ when being used in the black-box setting to explain the predictions of a SVR model.   
Values shown are RMSE averaged over 25 trials.  
Bold entries denote a significant difference.}

\end{minipage}

\vspace{5pt}
\begin{tabular}{|ccccc|}
\hline
\textbf{Dataset} & \textbf{n} & \textbf{p} & \textbf{d - RF} & \textbf{d - GBRT} \\ \hline
Autompgs         & 392        & 8          & 6.44            & 5.94              \\
Communities      & 1993       & 103        & 54.14           & 50.12             \\
Crimes           & 2214       & 103        & 20.34           & 21.62             \\
Day              & 731        & 15         & 2.46            & 3.02              \\
Happiness        & 578        & 8          & 7.74            & 7.46              \\
Housing          & 506        & 12         & 9.98            & 10.06             \\
Music            & 1059       & 70         & 5.56            & 14.46             \\
Winequality-red  & 1599       & 12         & 7.1             & 6.88              \\ \hline
\end{tabular}
\caption{\label{uci-features} \small
For each dataset, its size, $n$, and dimension, $p$, along with the average number of features used, $d$, by \name for RF and GBRT respectively.}

\end{table}

\subsection{Accuracy on UCI datasets} \label{accuracy}

We run our experiments on several of the UCI datasets \cite{Dua:2017}.  
Each dataset was divided into a 50/25/25 training, validation, and testing split for each of the trials.  
All variables, including the response, were standardized to have mean zero and variance one.

We compare \name and SILO with tree ensembles constructed using standard random forests (RF) and gradient boosted regression trees (GBRT).  
The ensemble choice for the baseline impacts the structure of the trees, which alters the weights as well as the global features selected by DStump. 
We include the performance of a linear model (LM) as well.  
Root mean squared errors (RMSE) are reported in Table \ref{uci-accuracy} and the number of selected features is in Table \ref{uci-features}.  
Overall, \name does at least as well as the tree ensembles and SILO, and often does better (the Music dataset being the sole exception).  

\subsection{Faithful Self-Explanations}\label{ssec:self-explain}

We next demonstrate that the local linear model that \name  uses to make its prediction doubles as an effective local explanation. 
The general data processing is the same as in the previous section, and we restrict our results to the random forest based version of \name. 
We use our proposed causal metric defined in \eqref{eq:causal} as our evaluation metric, defining $p_x$ as $\mathcal{N}(x, \sigma I)$, using the squared $l_2$ loss, and approximating the expectation by taking $x$ from the testing set and drawing five $x'$ per testing point. 

We chose $\sigma = 0.1$ as a reasonable choice for the neighborhood scale because the data was normalized to have variance one. 
The results, which show the RMSE of the causal metric in Table \ref{as-self},  demonstrate that the local linear models produced are good local explanations for the model as a whole when compared to using LIME to explain \name. 

\subsection{\name as a Black-box Explainer} \label{blackbox}

The overall setup for these experiments is the same as in the previous section, except that we are evaluating explanation systems in the black-box setting, where they are fit against the predictive model's predicted response. 
We use a Support Vector Regression (SVR) model (implementation and standard parameters from scikit-learn) as a black-box predictive model.
We present a comparison of \name to LIME  in Table \ref{as-bb}.  
Our results show the RMSE of the causal metric.  
\name produces more accurate local explanations than LIME for all but the Crimes dataset, where the difference was not statistically significant.

\textbf{What about a Larger $\bm{\sigma}$?}:
After data normalization, the median range of our features (across all datasets) is roughly six.  
For a single dimension, the width of the 95\% confidence interval for the sample used in the causal metric is approximately $4 \sigma$.  
In addition to the reported results with $\sigma=0.1$, we ran experiments with $\sigma = 0.25$.  
For this larger value, the expected range of the neighborhood is approximately one sixth of the overall feature range.
Further, the probability that one dimension falls outside of this interval increases exponentially with the number of dimensions.  
Consequently, the neighborhoods are even larger for high dimensional problems. 
Thus $\sigma = 0.25$ appears to be unreasonably large for a local explanation in high dimensions, especially considering that standard nearest neighbor methods  frequently rely on a small constant number of neighbors. 
Nonetheless, we note that for $\sigma = 0.25$, we found that \name significantly outperformed LIME on the Autompgs, Happiness, Housing, and Winequality-red datasets. 
In contrast, LIME significantly outperformed \name on the three datasets with largest dimension $p$ (see Table~\ref{uci-features}): Communities, Crimes, and Music. 

\subsection{Using Influential Training Points} \label{training-points}

To demonstrate how the local training distribution can be used to make inferences about the global patterns of the model, we work with the datasets introduced in Fig. \ref{fig:3datasets} of Sec.~\ref{sec:intro}. 
Each dataset consists of $n=200$ draws from $[0,1]^5$ taken uniformly at random.  
These samples are passed through either a linear, shifted inverse logistic (SIL), or step function that acts only on the first dimension of the feature vector (the remaining four dimensions are noisy features), and then a normally distributed noise is added with $\sigma = 0.1$.  
We refer to the first feature as the `active' feature.
We then fit a random forest to the data and fit \name to the random forest's predicted response. 
Next, we do a grid search across the active feature (e.g., $[x]_0 = 0, 0.1, \ldots, 1.0)$ and sample the remaining features uniformly at random from $[0,1]$ (simulating sampling the remaining features from the data distribution).  
For a given sampled point $x'$, we use \name's local training distribution to identify the 20 most influential training points
and plot a boxplot of the distribution of the active feature for these influential points.  
To smooth the results, we repeat this procedure 10 times for each point in the grid search. 
The plots of these distributions are in Fig. \ref{fig:distributions}. 

\textbf{Interpreting these Distributions: }
When a random forest fits a continuous function, each tree splits the input space into finer and finer partitions and, if the slope of the function does not change too rapidly, the distribution of where the partitions are split are relatively uncorrelated between the trees.  
As a result, the influential training points for a prediction at $x$ are roughly centered around $x$ and tend to change smoothly as $x$ changes.
This is demonstrated in Fig. \ref{fig:distributions}a with the exception of points near the boundary of the distribution.  

Further, the steeper the function is at $x$, the finer the partitions become around $x$ and, consequently, the distribution of the influential points becomes more concentrated.  
This can be seen by contrasting Fig. \ref{fig:distributions}a and Fig. \ref{fig:distributions}b; in Fig. \ref{fig:distributions}a, the distributions have roughly equal variance while, in Fig. \ref{fig:distributions}b, the distributions in the flat areas of the function have large variance while the distributions in the steep transition of the function are much narrower. 
Although not shown, one extreme of this is a feature that is inactive;  in all of our experiments, the inactive feature distributions of the most influential training points look like the original data distribution.  

In the extreme case, when fitting a discontinuous function, the trees are likely to all split at or near the discontinuity.  
As a result, influential training points for $x$ may not be centered around $x$ when $x$ is near the discontinuity and they will change abruptly as $x$ moves past the discontinuity.  
This is demonstrated in Fig. \ref{fig:distributions}c where we clearly observe the transitions of the step function, as well as the fact that the intervals are not centered around $x$.  
Further, in Fig. \ref{fig:distributions}b we can see the fact that there are two flat areas of the function with a steep and short transition between by noticing that there are two main ranges for the influential points with one intermediate range at $x_0 = 0.5$.  
 
\section{Conclusion and Future Work}
We have shown that \name is effective both as a predictive model and an explanation system.  
Additionally, we have demonstrated how to use its local training distribution to address two key weaknesses of local explanations:  
1) Detecting and modeling global patterns, and 
2) Determining whether an exemplar explanation can be applied to a new test point. 
Some interesting avenues of future work include: 
1) Exploiting the fact that  \name is a locally linear model to tap into the wide range of approaches that use influence functions to improve model accuracy or identify interesting data points via measures such as leverage and Cook's distance \cite{chatterjee1986influential};  
2) Exploring the use of local feature selection approaches with \name, e.g., by considering impurity reductions along the paths through all trees for a given test point; and 
3) Exploring methods other than tree ensembles for defining the similarity weights from Eq. \ref{weight}.

\section*{Acknowledgements}
This work was supported in part by DARPA FA875017C0141, the National Science Foundation grants IIS1705121 and IIS1838017, an Okawa Grant, a Google Faculty Award, an Amazon Web Services Award, and a Carnegie Bosch Institute Research Award. Any opinions, findings and conclusions or recommendations expressed in this material are those of the author(s) and do not necessarily reflect the views of DARPA, the National Science Foundation, or any other funding agency.

\bibliographystyle{plain}
\bibliography{bibliography}

\end{document}